\documentclass{article}

\usepackage{times}
\usepackage{graphicx}
\usepackage{subfigure}

\usepackage{natbib}

\usepackage{algorithm}
\usepackage{algorithmic}

\usepackage{hyperref}

\usepackage[accepted]{icml2016}

\usepackage{amsmath}
\usepackage{amssymb}
\usepackage{amsfonts}
\usepackage{color}
\usepackage{capt-of}

\icmltitlerunning{Associative Long Short-Term Memory}

\newcommand{\elemmul}{\odot}
\newcommand{\complexmul}{\circledast}

\newcommand{\sigmoid}{\sigma}
\newcommand{\bound}{\mathrm{bound}}
\newcommand{\fig}[1]{Figure~\ref{fig:#1}}
\newcommand{\target}[1]{\underline{#1}}
\newcommand{\figwidth}{0.41\textwidth}
\newcommand{\figskip}{\vskip -0.2cm}
\newcommand{\eqcomment}[1]{}
\newcommand*\conj[1]{\overline{#1}}

\newcommand{\onlysupplement}[1]{#1}

\begin{document}

\onlysupplement{
\twocolumn[
\icmltitle{Associative Long Short-Term Memory}

\icmlauthor{Ivo Danihelka}{danihelka@google.com}
\icmlauthor{Greg Wayne}{gregwayne@google.com}
\icmlauthor{Benigno Uria}{buria@google.com}
\icmlauthor{Nal Kalchbrenner}{nalk@google.com}
\icmlauthor{Alex Graves}{gravesa@google.com}
\icmladdress{Google DeepMind}

\icmlkeywords{LSTM, ICML}

\vskip 0.3in
]

\begin{abstract}
We investigate a new method to augment recurrent neural networks with extra memory without increasing the number of network parameters.
The system has an associative memory based on complex-valued vectors and is closely related to Holographic Reduced Representations and Long Short-Term Memory networks. Holographic Reduced Representations have limited capacity: as they store more information, each retrieval becomes noisier due to interference. 
Our system in contrast creates redundant copies of stored information, which enables retrieval with reduced noise. 
Experiments demonstrate faster learning on multiple memorization tasks.
\end{abstract}

\section{Introduction}

We aim to enhance LSTM \citep{hochreiter1997long}, which in recent years has become widely used in sequence prediction, speech recognition and machine translation \citep{graves2013generating,graves2013speech,sutskever2014sequence}.
We address two limitations of LSTM. The first limitation is that the number of memory cells is linked to the size of the recurrent weight matrices.
An LSTM with $N_h$ memory cells requires a recurrent weight matrix with $O(N^2_h)$ weights. 
The second limitation is that LSTM is a poor candidate for learning to represent common data structures like arrays because it lacks a mechanism to index its memory while writing and reading.

To overcome these limitations,
recurrent neural networks have been previously augmented with
soft or hard attention mechanisms to external memories \citep{graves2014ntm,sukhbaatar2015end,stackrnn,grefenstette2015learning,zaremba2015reinforcement}.
The attention acts as an addressing system that selects memory locations. 
The content of the selected memory locations can be read or modified by the network.

We provide a different addressing mechanism in Associative LSTM, where, like LSTM, an item is stored in a distributed vector representation without locations.
Our system implements an associative array that stores key-value pairs based on two contributions:
\begin{enumerate}
  \item We combine LSTM with ideas from Holographic Reduced Representations (HRRs) \citep{plate2003holographic} to enable key-value storage of data.
  \item A direct application of the HRR idea leads to very lossy storage. We use redundant storage to increase memory capacity and to reduce noise in memory retrieval.
\end{enumerate}

HRRs use a ``binding'' operator to implement key-value binding between two vectors (the key and its associated content). They natively implement associative arrays; as a byproduct, they can also easily implement stacks, queues, or lists. Since Holographic Reduced Representations may be unfamiliar to many readers, Section~\ref{sec:representing_seqs} provides a short introduction to them and to related vector-symbolic architectures \citep{kanerva2009hyperdimensional}.

In computing, Redundant Arrays of Inexpensive Disks (RAID) provided a means to build reliable storage from unreliable components. 
We similarly reduce retrieval error inside a holographic representation by using redundant storage, a construction described in Section~\ref{sec:redundantMemory}. 
We then combine the redundant associative memory with LSTM in Section~\ref{sec:redundantLSTM}. The system can be equipped with a large memory without increasing the number of network parameters. Our experiments in Section~\ref{sec:experiments} show the benefits of the memory system for learning speed and accuracy.

\section{Background}
\label{sec:representing_seqs}
Holographic Reduced Representations are a simple mechanism to represent an associative array of key-value pairs in a fixed-size vector. 
Each individual key-value pair is the same size as the entire associative array; the array is represented by the sum of the pairs. Concretely, consider a complex vector key $r=(a_r[1] e^{i \phi_r[1]}, a_r[2] e^{i \phi_r[2]}, \dots)$, which is the same size as the complex vector value $x$.
The pair is ``bound'' together by element-wise complex multiplication, which multiplies the moduli and adds the phases of the elements:
\begin{align}
\label{eq:conv}
y &= r \complexmul x \\
  &= (a_r[1] a_x[1] e^{i (\phi_r[1] + \phi_x[1])}, a_r[2] a_x[2] e^{i (\phi_r[2] + \phi_x[2])}, \dots)
\end{align}

Given keys $r_1, r_2, r_3$ and input vectors $x_1, x_2, x_3$, the associative array is
\begin{align}
 c = r_1 \complexmul x_1 + r_2 \complexmul x_2 + r_3 \complexmul x_3
\end{align}
where we call $c$ a \textit{memory trace}.

Define the \emph{key inverse}:
\begin{align}
  r^{-1} = (a_r[1]^{-1} e^{-i \phi_r[1]}, a_r[2]^{-1} e^{-i \phi_r[2]}, \dots)
\end{align} 
To retrieve the item associated with key $r_k$, we multiply the memory trace element-wise by the vector $r_k^{-1}$. For example:
\begin{align}
	r_2^{-1} \complexmul c \nonumber &= r_2^{-1} \complexmul (r_1 \complexmul x_1 +  r_2 \complexmul x_2 + r_3
  \complexmul x_3 ) \nonumber \\
  &= x_2 + r_2^{-1} \complexmul (r_1 \complexmul x_1 + r_3 \complexmul x_3) \nonumber \\
  & = x_2 + \mathit{noise}
\end{align}
The product is exactly $x_2$ together with a noise term.
If the phases of the elements of the key vector are randomly distributed, the noise term has zero mean.

Instead of using the key inverse, Plate recommends using the complex conjugate of the key, $\conj{r_k} = (a_k[1] e^{-i \phi_k[1]}, a_k[2] e^{-i \phi_k[2]}, \dots)$, for retrieval: the elements of the exact inverse have moduli $(a_k[1]^{-1}, a_k[2]^{-1}, \dots)$, which can magnify the noise term. Plate presents two different variants of Holographic Reduced Representations. The first operates in real space and uses circular convolution for binding; the second operates in complex space and uses element-wise complex multiplication. The two are related by Fourier transformation. See \cite{plate2003holographic} or \cite{kanerva2009hyperdimensional} for a more comprehensive overview.

\section{Redundant Associative Memory}
\label{sec:redundantMemory}

\begin{figure*}[t]
\begin{center}
\includegraphics[width=\linewidth]{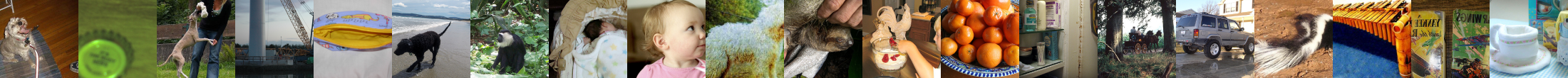}
\includegraphics[width=\linewidth]{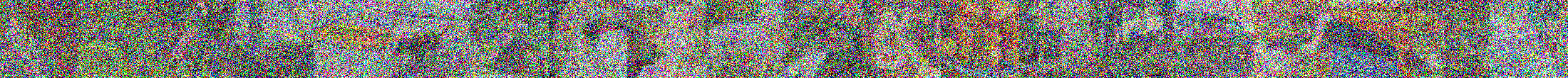}
\includegraphics[width=\linewidth]{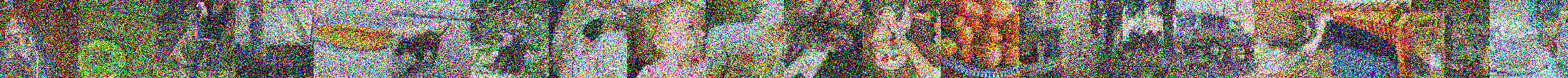}
\includegraphics[width=\linewidth]{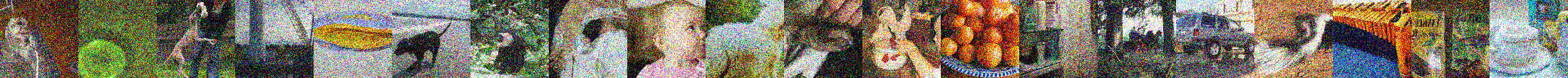}
\end{center}
\caption{From top to bottom: 20 original images and the image sequence retrieved from 1 copy, 4 copies, and 20 copies of the memory trace. Using more copies reduces the noise.
}
\label{fig:restoredImgs}
\end{figure*}

\begin{figure*}[t]
\begin{center}
\includegraphics[width=0.33\linewidth]{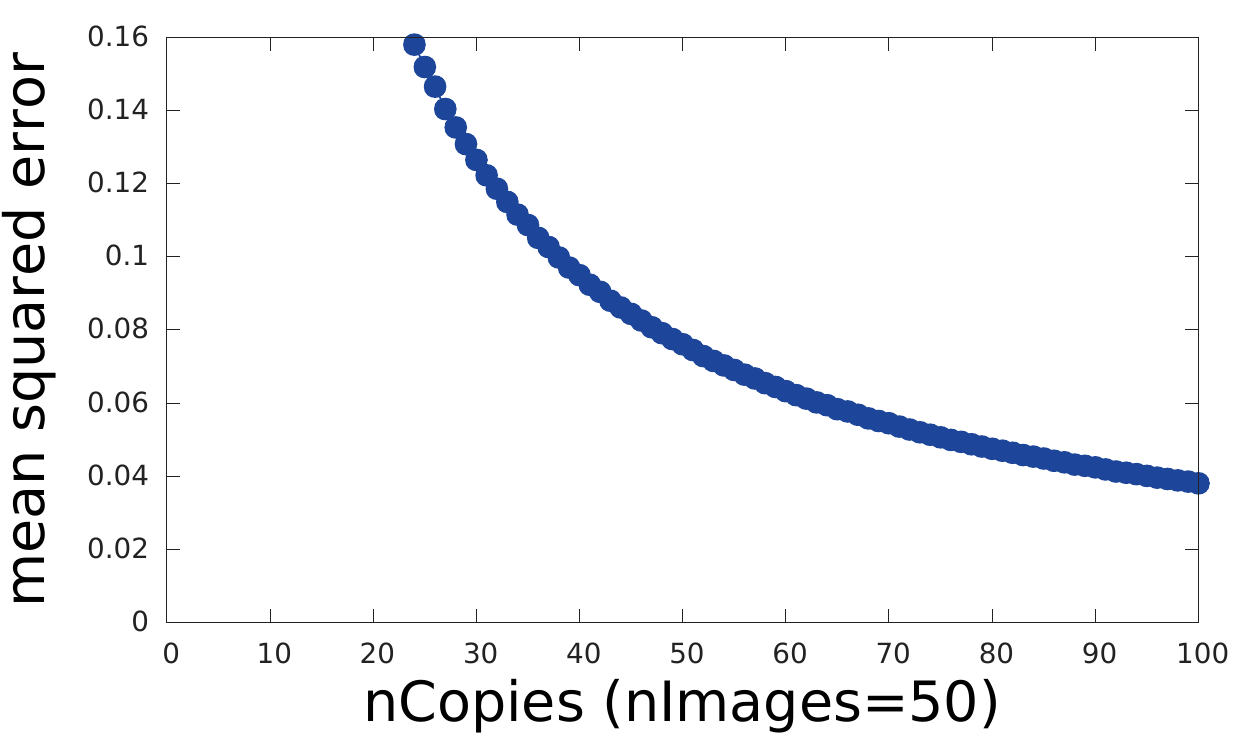}
\includegraphics[width=0.33\linewidth]{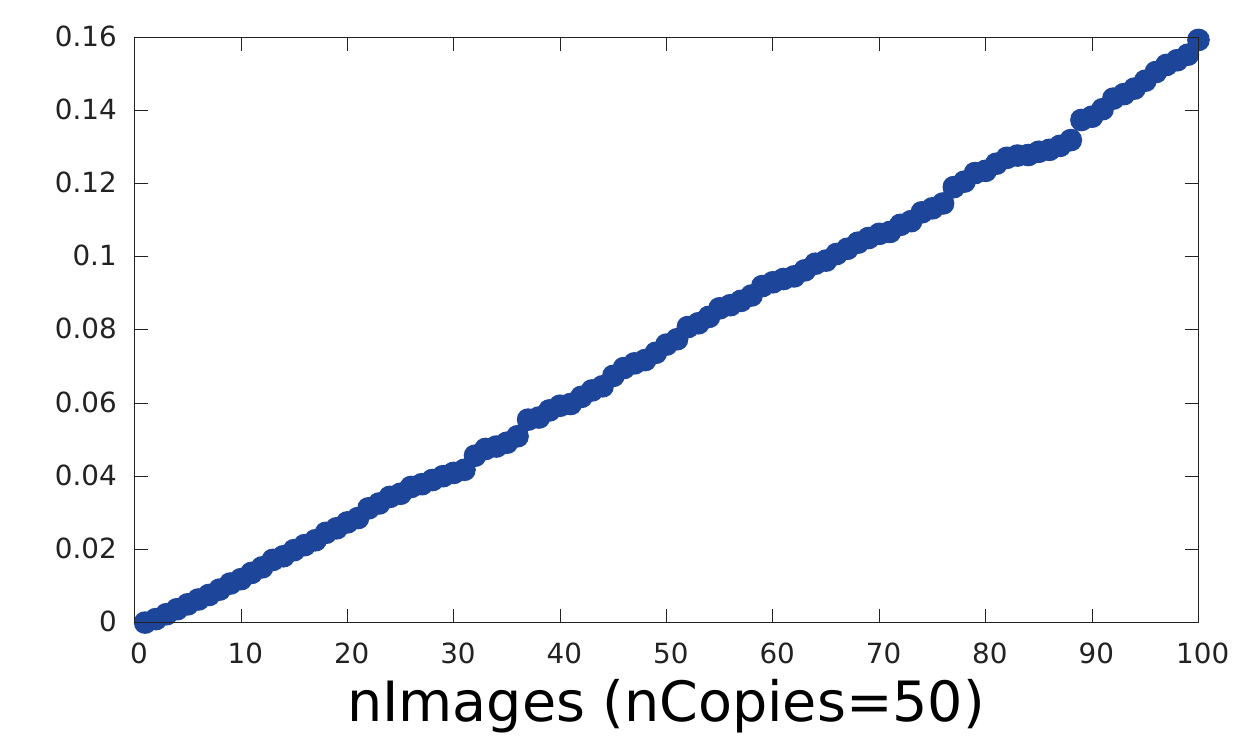}
\includegraphics[width=0.33\linewidth]{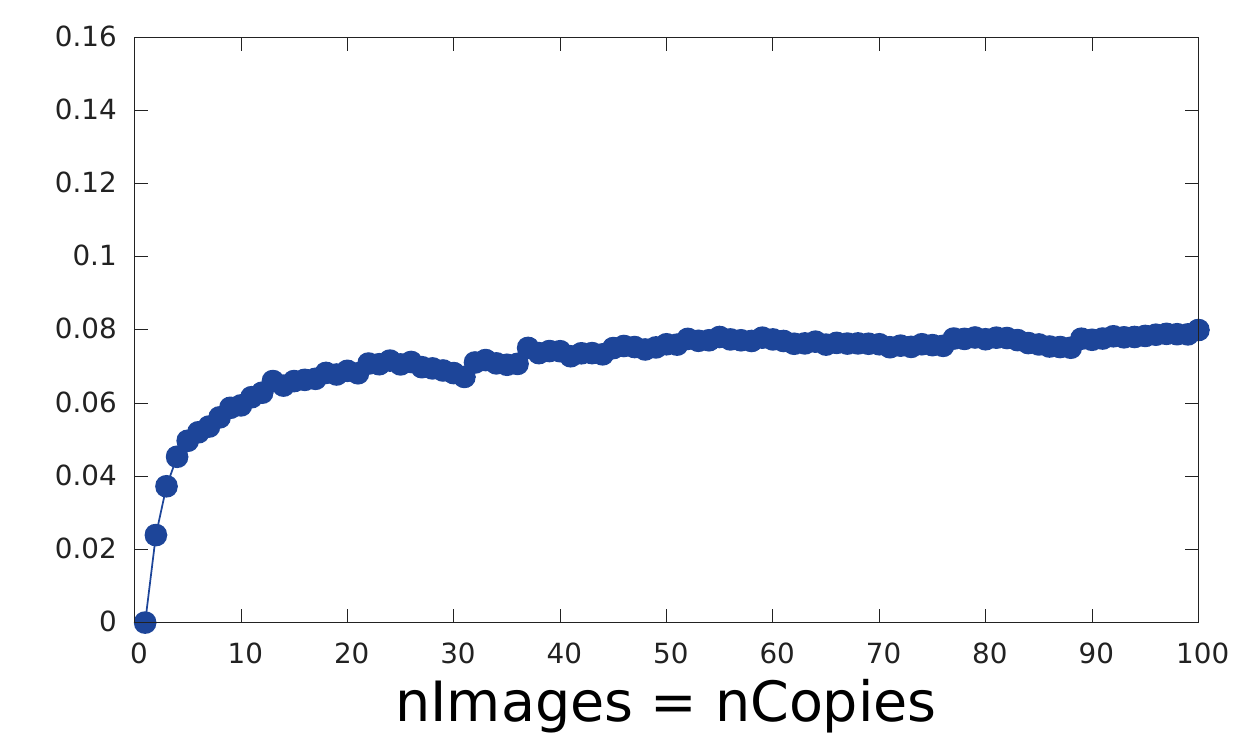}
\end{center}
\caption{The mean squared error per pixel when retrieving an ImageNet image from a memory trace.
\textbf{Left:} 50 images are stored in the memory trace. The number of copies ranges from 1 to 100.
\textbf{Middle:} 50 copies are used, and the number of stored images goes from 1 to 100. The mean squared error grows linearly.
\textbf{Right:} The number of copies is increased together with the number of stored images. After reaching 50 copies, the mean squared error is almost constant.
}
\label{fig:mse}
\end{figure*}

As the number of items in the associated memory grows, the noise incurred during retrieval grows as well.
Here, we propose a way to reduce the noise by storing multiple transformed copies of each input vector. When retrieving an item, we compute the average
of the restored copies. 
Unlike other memory architectures \citep{graves2014ntm,sukhbaatar2015end}, the memory does not have a discrete number of memory locations;
instead, the memory has a discrete number of copies.

Formally, let $c_s$ be the memory trace for the \mbox{$s$-th} copy:
\begin{align}
  \label{eq:copyTrace}
  c_s &= \sum_{k=1}^{N_\mathit{items}} (P_s r_k) \complexmul x_k
\end{align}
where $x_k \in \mathbb{C}^{N_h/2}$ is the \mbox{$k$-th} input; $r_k \in \mathbb{C}^{N_h/2}$ is the key. Each complex number contributes a real and imaginary part, so the input and key are represented with $N_h$ real values. $P_s \in \mathbb{R}^{N_h/2 \times N_h/2}$ is a constant random permutation, specific to each copy. Permuting the key decorrelates the retrieval noise from each copy of the memory trace. 

When retrieving the \mbox{$k$-th} item, we compute the average over all copies:
\begin{align}
  \label{eq:copyAvg}
  \tilde{x}_k &= \frac{1}{N_\mathit{copies}}\sum^{N_\mathit{copies}}_{s=1} \conj{(P_s r_k)} \complexmul c_s
\end{align}
where $\conj{(P_s r_k)}$ is the complex conjugate of $P_s r_k$.

Let us examine how the redundant copies reduce retrieval noise by inspecting a retrieved item.
If each complex number in $r_k$ has modulus equal to 1,
its complex conjugate acts as an inverse, and the retrieved item is:
\begin{align}
  \tilde{x}_k &= x_k +  \frac{1}{N_\mathit{copies}}\sum^{N_\mathit{copies}}_{s=1} \sum^{N_\mathit{items}}_{\substack{k' \neq k}} \conj{(P_s r_k)} \complexmul (P_s r_{k'}) \complexmul x_{k'} \nonumber \\
    &= x_k +  \sum^{N_\mathit{items}}_{k' \neq k}  x_{k'}  \complexmul \frac{1}{N_\mathit{copies}}\sum^{N_\mathit{copies}}_{s=1} P_s (\conj{r_k} \complexmul r_{k'}) \nonumber \\
    &= x_k + \mathit{noise}
\end{align}
where the $\sum^{N_\mathit{items}}_{k' \neq k}$ sum is over all other stored items. If the terms $P_s (\conj{r}_k \complexmul r_{k'})$ are independent, they add incoherently when summed over the copies. Furthermore, if the noise due to one item $\xi_{k'} = x_{k'}  \complexmul \frac{1}{N_\mathit{copies}}\sum^{N_\mathit{copies}}_{s=1} P_s (\conj{r_k} \complexmul r_{k'})$ is independent of the noise due to another item, then the variance of the total noise is $O(\frac{N_\mathit{items}}{N_\mathit{copies}})$. Thus, we expect that the retrieval error will be roughly constant if the number of copies scales with the number of items.

Practically, we demonstrate that the redundant copies with random permutations are effective at reducing retrieval noise in Figure~\ref{fig:restoredImgs}. We take a sequence of ImageNet images \citep{ILSVRC15}, each of dimension $3 \times 110 \times 110$, where the first dimension is a colour channel.
We vectorise each image and consider the first half of the vector to be the real part and the second half the imaginary part of a complex vector.
The sequence of images is encoded by using random keys with moduli equal to 1.
We see that retrieval is less corrupted by noise as the number of copies grows. 

The mean squared error of retrieved ImageNet images is analysed in Figure~\ref{fig:mse} with varying numbers of copies and images.
The simulation agrees accurately with our prediction:
the mean squared error is proportional to the number of stored items and inversely proportional to the number of copies.

The redundant associative memory has several nice properties:
\begin{itemize}
\item The number of copies can be modified at any time: reducing their number increases retrieval noise, while increasing the number of copies enlarges capacity.
\item It is possible to query using a partially known key by setting some key elements to zero.
Each copy of the permuted key $P_s r_k$ routes the zeroed key elements to different dimensions.
We need to know $O(\frac{N_h}{N_\mathit{copies}})$ elements of the key to recover the whole value.
\item Unlike Neural Turing Machines \citep{graves2014ntm}, it is not necessary to search for free locations when writing.
\item It is possible to store more items than the number of copies at the cost of increased retrieval noise.
\end{itemize}

\section{Long Short-Term Memory}
We briefly introduce the LSTM with forget gates \citep{gers2000learning}, a recurrent neural network whose hidden state is described by two parts $h_{t}, c_{t} \in \mathbb{R}^{N_h}$. At each time step, the network is presented with an input $x_t$ and updates its state to
\begin{align}
  \hat{g}_f, \hat{g}_i, \hat{g}_o, \hat{u} &= W_{xh} x_t + W_{hh} h_{t-1} + b_{h} \\
   g_f &= \sigmoid(\hat{g}_f) \eqcomment{forget gate} \\
  g_i &= \sigmoid(\hat{g}_i) \eqcomment{input gate} \\
  g_o &= \sigmoid(\hat{g}_o) \eqcomment{output gate} \\
  u &= \tanh(\hat{u}) \eqcomment{update} \\
  c_t &= g_f \elemmul c_{t-1} + g_i \elemmul u \eqcomment{cell state} \\
  h_t &= g_o \elemmul \tanh(c_t)
\end{align}

where $\sigmoid(x) \in (0,1)$ is the logistic sigmoid function, and $g_f, g_i, g_o$
are the \textit{forget gate}, \textit{input gate}, and \textit{output gate}, respectively.
The $u$ vector is a proposed \textit{update} to the \textit{cell} state $c$.
$W_{xh}, W_{hh}$ are weight matrices, and $b_{h}$ is a bias vector.
``$\elemmul$'' denotes element-wise multiplication of two vectors.

\section{Associative Long Short-Term Memory}
\label{sec:redundantLSTM}
When we combine Holographic Reduced Representations with the LSTM, we need to implement complex vector multiplications. For a complex vector $z = h_\textit{real} + i h_\textit{imaginary}$, we use the form
\begin{align}
  h &= \left[\begin{array}{l}
    h_\textit{real} \\
    h_\textit{imaginary} \\
  \end{array}\right]
\end{align}
where $h \in \mathbb{R}^{N_h}$, $h_\textit{real}, h_\textit{imaginary} \in \mathbb{R}^{N_h/2}$. In the network description below, the reader can assume that all vectors and matrices are strictly real-valued.

As in LSTM, we first compute gate variables, but we also produce parameters that will be used to define associative keys $\hat{r}_{i}, \hat{r}_{o}$. The same gates are applied to the real and imaginary parts:
\begin{align}
\hat{g}_f, \hat{g}_i, \hat{g}_o, \hat{r}_i, \hat{r}_o &= W_{xh} x_t + W_{hh} h_{t-1} + b_{h} \\
  \label{eq:whu}
  \hat{u} &= W_{xu} x_t + W_{hu} h_{t-1} + b_{u} \\
  g_f &= \left[\begin{array}{l}
    \sigmoid(\hat{g}_f) \\
    \sigmoid(\hat{g}_f) \\
  \end{array}\right] \eqcomment{forget gate} \\
  g_i &= \left[\begin{array}{l}
    \sigmoid(\hat{g}_i) \\
    \sigmoid(\hat{g}_i) \\
  \end{array}\right]  \eqcomment{input gate} \\
  g_o &= \left[\begin{array}{l}
    \sigmoid(\hat{g}_o) \\
    \sigmoid(\hat{g}_o) \\
  \end{array}\right] \eqcomment{output gate}
 \end{align} 
 
Unlike LSTM, we use an activation function that operates only on the modulus of a complex number. The following function restricts the modulus of a $(\textit{real}, \textit{imaginary})$ pair to be between 0 and 1:
\begin{align}
\bound(h) &= \left[\begin{array}{l}
    h_\textit{real} \oslash d \\
    h_\textit{imaginary} \oslash d \\
  \end{array}\right]
\end{align}
where ``$\oslash$'' is element-wise division, and $d \in \mathbb{R}^{N_h / 2} = \max(1, \sqrt{h_\textit{real} \elemmul h_\textit{real} + h_\textit{imaginary} \elemmul h_\textit{imaginary}})$ corresponds to element-wise normalisation by the modulus of each complex number when the modulus is greater than one.
This hard bounding worked slightly better than applying $\tanh$ to the modulus. ``$\bound$'' is then used to construct the \emph{update} and two keys:
 \begin{align}
  u &= \bound(\hat{u})  \eqcomment{update} \\
  r_i &= \bound(\hat{r}_i) \eqcomment{input key} \\
  r_o&= \bound(\hat{r}_o) \eqcomment{output key} 
\end{align}
where $r_i \in \mathbb{R}^{N_h}$ is an \textit{input key}, acting as a storage key in the associative array,
and $r_o \in \mathbb{R}^{N_h}$ is an \textit{output key}, corresponding to a lookup key.
The update $u$ is multiplied with the input gate $g_i$
to produce the value to be stored.

Now, we introduce redundant storage and provide the procedure for memory retrieval.
For each copy, indexed by $s \in \{1, \dots, N_\mathit{copies}\}$, we add the same key-value pair to the cell state:
\begin{align}
  \eqcomment{permuted keys:}
  r_{i,s} &= \left[\begin{array}{cc}
    P_s & 0 \\
    0 & P_s \\
  \end{array}\right] r_i
  \end{align}
  \begin{align}
   \eqcomment{cell state:}
   \label{eq:cellupdate}
  c_{s,t} &= g_f \elemmul c_{s,t-1} + r_{i,s} \complexmul (g_i \elemmul u)
  \end{align} 
where $r_{i,s}$ is the permuted input key; $P_s \in \mathbb{R}^{N_h/2 \times N_h/2}$ is a constant random permutation matrix, specific to the \mbox{$s$-th} copy. ``$\complexmul$'' is element-wise complex multiplication and is computed using
\begin{align}
  r \complexmul u &= \left[\begin{array}{l}
  r_\textit{real} \elemmul u_\textit{real} - r_\textit{imaginary} \elemmul u_\textit{imaginary} \\
  r_\textit{real} \elemmul u_\textit{imaginary} + r_\textit{imaginary} \elemmul u_\textit{real} \\
  \end{array}\right]
\end{align}

The output key for each copy, $r_{o,s}$, is permuted by the same matrix as the copy's input key:
  \begin{align}
  r_{o,s} &= \left[\begin{array}{cc}
    P_s & 0 \\
    0 & P_s \\
  \end{array}\right] r_o 
  \end{align}
Finally, the cells (memory trace) are read out by averaging the copies:
 \begin{align}
 h_t &= g_o \elemmul \bound\bigg(\frac{1}{N_\mathit{copies}} \sum^{N_\mathit{copies}}_{s=1} r_{o,s} \complexmul c_{s,t}\bigg)
\end{align}

Note that permutation can be performed in $O(N_h)$ computations.
Additionally, all copies can be updated in parallel by operating on tensors of size $N_\mathit{copies} \times N_h$.

On some tasks, we found that learning speed was improved by not feeding $h_{t-1}$ to the update $u$: namely, $W_{hu}$ is set to zero in Equation~\ref{eq:whu}, which causes $u$ to serve as an embedding of $x_t$. This modification was made for the episodic copy, XML modeling, and variable assignment tasks below.

\section{Experiments}
\label{sec:experiments}

All experiments used the Adam optimisation algorithm \citep{kingma2014adam} with no gradient clipping.
For experiments with synthetic data, we generate new data for each training minibatch, obviating the need for a separate test data set. Minibatches of size 2 were used in all tasks beside the Wikipedia task below, where the minibatch size was 10.

\begin{table*}[t]
\caption{Network sizes on the episodic copy task.}
\label{tab:netdetails}
\begin{center}
\begin{tabular}{l l l r}
\hline
Network & $N_h$ & Relative speed & \#parameters \\
\hline
LSTM & $128$ & 1 & $72,993$ \\
LSTM nH=512 & $512$ & 0.18 & $1,078,305$ \\
\hline
Associative LSTM & $128$ (=$64$ complex numbers) & \begin{tabular}{@{}l@{}}
  0.22 ($N_\mathit{copies}=1$) \\
  0.16 ($N_\mathit{copies}=4$) \\
  0.12 ($N_\mathit{copies}=8$)
\end{tabular} & $65,505$ \\
\hline
Permutation RNN & $256$ & 2.05 & $5,642$ \\
Unitary RNN & $256$ (=$128$ complex numbers) & 0.24 & $6,666$ \\
Multiplicative uRNN & $256$ & 0.23 & $10,506$ \\
\hline
\end{tabular}
\end{center}
\end{table*}

We compared Associative LSTM to multiple baselines:

\textbf{LSTM.}
We use LSTM with forget gates and without peepholes \citep{gers2000learning}.

\textbf{Permutation RNN.}
Each sequence is encoded by using powers of a constant random permutation matrix as keys:
\begin{align}
  h_t &= P h_{t-1} + W x_t
\end{align}
Only the input and output weights are learned.
Representing sequences by ``permuting sums''
is described in \cite{kanerva2009hyperdimensional}.

\textbf{Unitary RNN.}
\cite{urnn} recently introduced recurrent neural networks
with unitary weight matrices.\footnote{We were excited to see that other researchers are
also studying the benefits of complex-valued recurrent networks.}
They consider dynamics of the form
\begin{align}
   h_t &= f(W h_{t-1} + V x_t)
\end{align}
where $W$ is a unitary matrix ($W^\dagger W = I$). The product of unitary matrices is a unitary matrix, so $W$ can be parameterised as the product of simpler unitary matrices. In particular,    
\begin{align}
   h_t &= f(D_3 R_2 F^{-1} D_2 P R_1 F D_1 h_{t-1} + V x_t)
\end{align}
where $D_3, D_2, D_1$ are learned diagonal complex matrices, and $R_2, R_1$ are learned reflection matrices. Matrices $F$ and $F^{-1}$ are the discrete Fourier transformation and its inverse.
$P$ is any constant random permutation. The activation function $f(h)$ applies a rectified linear unit with a learned bias to the modulus of each complex number. Only the diagonal and reflection matrices, $D$ and $R$, are learned, so Unitary RNNs have fewer parameters than LSTM with comparable numbers of hidden units. 

\textbf{Multiplicative Unitary RNN.}
To obtain a stronger baseline, we enhanced the Unitary RNNs with multiplicative interactions \citep{sutskever2011generating} by conditioning all complex diagonal matrices on the input $x_t$:
\begin{align}
  \hat{r} &=  W_{xr} x_t \\
  D &= \left[\begin{array}{lr}
  \mathrm{diag}(\cos(\hat{r})) & -\mathrm{diag}(\sin(\hat{r})) \\
  \mathrm{diag}(\sin(\hat{r})) & \mathrm{diag}(\cos(\hat{r})) \\
  \end{array}\right]
\end{align}

\subsection{Episodic Tasks}
\subsubsection{Episodic Copy}
The copy task is a simple benchmark that tests the ability of the architectures to store a sequence of random characters and repeat the sequence after a time lag.
Each input sequence is composed of 10 random characters, followed by 100 blanks, and a delimiter symbol.
After the delimiter symbol is presented, networks must reproduce the first 10 characters, matching the task description in \cite{urnn}. Although copy is not interesting \emph{per se}, failure on copy indicates an extreme limitation of a system's capacity to memorise.

\begin{figure}[t]
\begin{center}
\includegraphics[width=\figwidth]{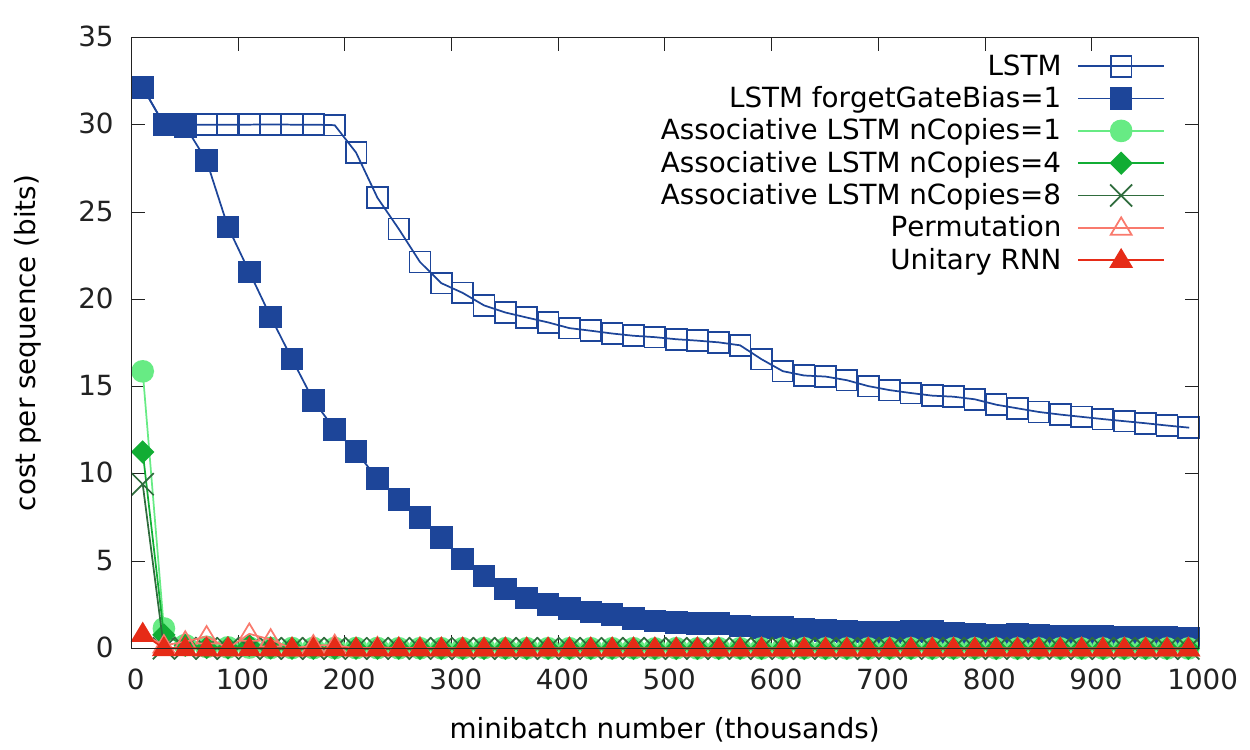}
\end{center}
\caption{Training cost per sequence on the fixed-length episodic copy task.
LSTM learns faster if the forget gate bias is set to 1. Associative LSTM was able to solve the task quickly without biasing the forget gate.
}
\label{fig:plot_episodic_copy}
\figskip
\end{figure}

\begin{figure}[t]
\begin{center}
\includegraphics[width=\figwidth]{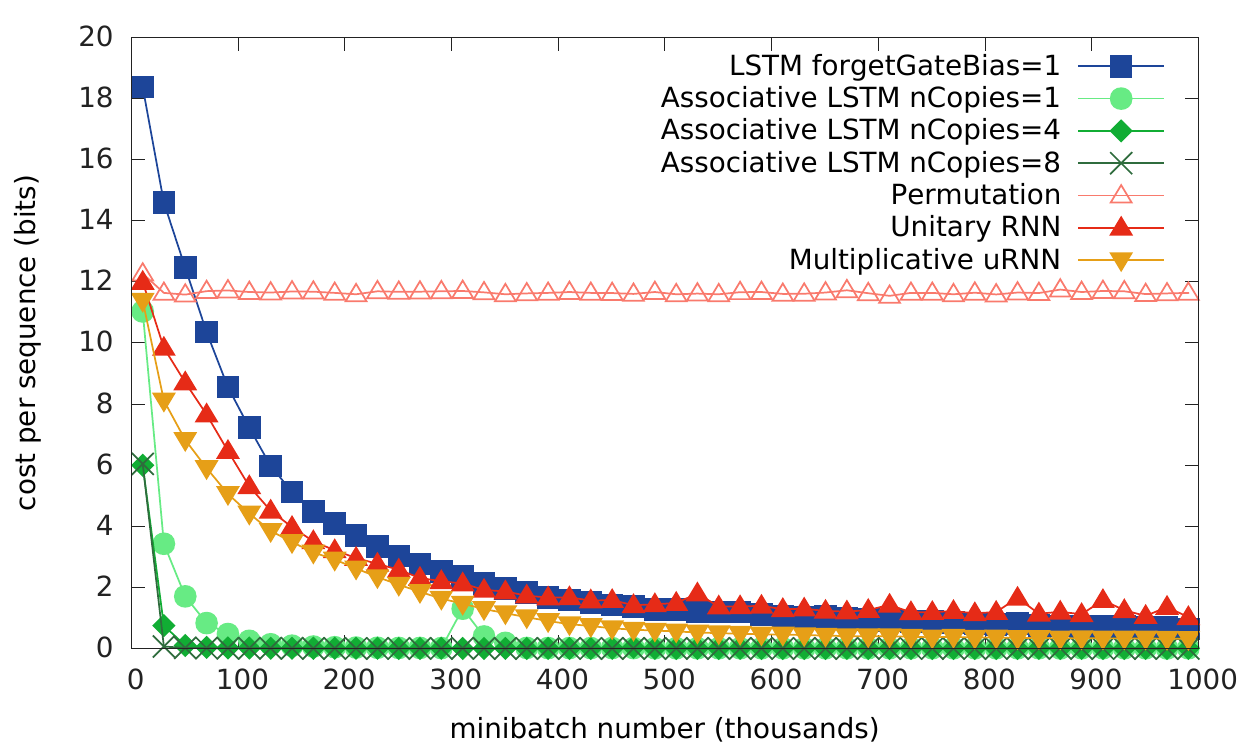}
\end{center}
\caption{Training cost per sequence on the episodic copy task with variable-length sequences (1 to 10 characters). Associative LSTM learns quickly and almost as fast as in the fixed-length episodic copy. Unitary RNN converges slowly relative to the fixed-length task.
}
\label{fig:plot_episodic_copy2}
\figskip
\end{figure}

\cite{urnn} presented very good results on the copy task using Unitary RNNs.
We wanted to determine whether Associative LSTM can learn the task with a similarly small number of data samples.
The results are displayed in Figure~\ref{fig:plot_episodic_copy}.
The Permutation RNN and Unitary RNN solve the task quickly.
Associative LSTM solved the task a little bit slower, but still much faster than LSTM.
All considered, this task requires the network to store only a small number of symbols; consequently, adding redundancy to the Associative LSTM, though not harmful, did not bestow any benefits.

We considered this variant of the copy task too easy since it posed no difficulty to the very simple Permutation RNN.
The Permutation RNN can find a solution by building a hash of the input sequence (a sum of many permuted inputs). 
The output weights then only need to learn to classify the hash codes. A more powerful Permutation RNN could use a deep network at the output.

To present a more complex challenge, we constructed one other variant of the task in which the number of random characters in the sequence is not fixed at 10 but is itself a variable drawn uniformly from 1 to 10. Surprisingly, this minor variation compromised the performance of the Unitary RNN, while the Associative LSTM still solved the task quickly. We display these results in Figure~\ref{fig:plot_episodic_copy2}.
We suspect that the Permutation RNN and Unitary RNN would improve on the task if they were outfitted with a mechanism to control the speed of the dynamics: for example, one could define a ``pause gate'' whose activation freezes the hidden state of the system after the first 10 symbols, including possible blanks. This would render the variable-length task exactly equivalent to the original.

Table~\ref{tab:netdetails} shows the number of parameters for each network.
Associative LSTM has fewer parameters than LSTM if the $W_{hu}$ matrix in Equation~\ref{eq:whu} is set to zero and the gates are duplicated for the real and the imaginary parts. Additionally, the number of parameters in Associative LSTM is not affected by the number of copies used; the permutation matrices do not add parameters since they are randomly initialised and left unchanged. 

\subsection{Online Tasks}
As important as remembering is forgetting. The following tasks consist of continuous streams of characters, and the networks must discover opportune moments to reset their internal states.

\begin{figure}[t]
\begin{center}
\includegraphics[width=\figwidth]{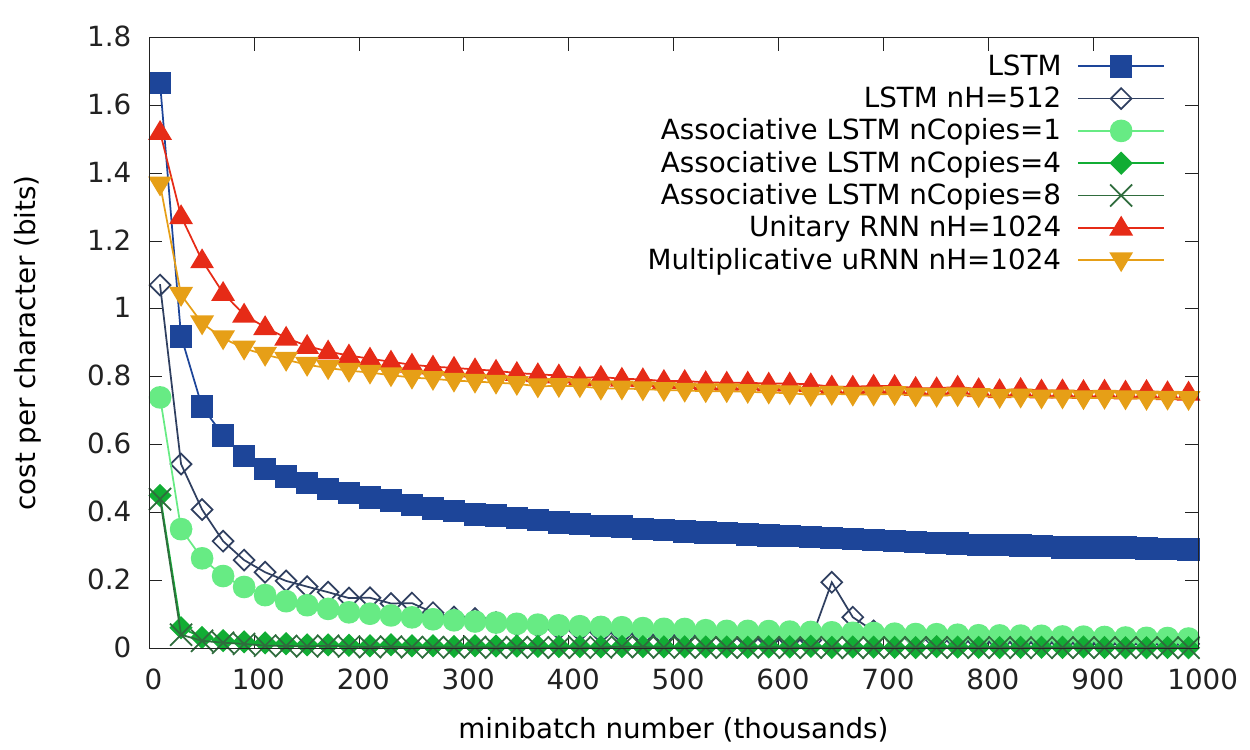}
\end{center}
\caption{Training cost on the XML task, including Unitary RNNs.
}
\label{fig:plot_xml_unitary}
\figskip
\end{figure}

\subsubsection{XML Modeling}

\begin{figure}[t]
\begin{center}
\includegraphics[width=\figwidth]{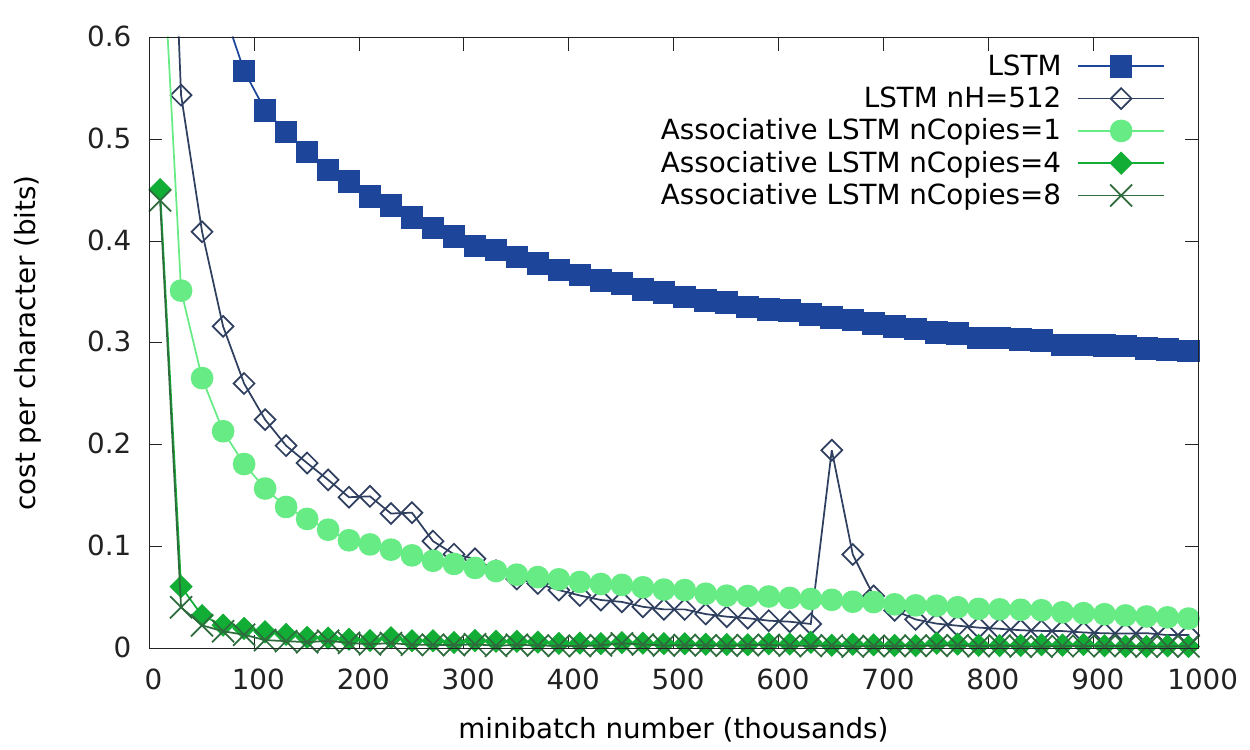}
\end{center}
\caption{Training cost on the XML task. LSTM and Associative LSTM with $128$ hidden units are also compared to a larger LSTM with $512$ units.}
\label{fig:plot_xml}
\figskip
\end{figure}

The XML modeling task was defined in \cite{jozefowicz2015empirical}. The XML language consists of nested (context-free) tags of the form ``\texttt{<tag1><tag2> \dots </tag2></tag1>}''.
The input is a sequence of tags with names of 1 to 10 random lowercase characters. The tag name is only predictable when it is closed by ``\texttt{</tag>}'', so the prediction cost is confined to the closing tags in the sequence. Each symbol must be predicted one time step before it is presented.
An example sequence looks like:

\texttt{\target{<}xkw>\target{<}svgquspn>\target{<}oqrwxsln>\target{<}/\target{oqrwxsln>}\target{<}/
\target{svgquspn>}\target{<}jrcfcacaa>\target{<}/\target{jrcfcacaa}>\target{<}/\target{xk}}\dots

with the cost measured only on the underlined segments. The XML was limited to a maximum depth of 4 nested tags to prevent extremely long sequences of opened tags when generating data. All models were trained with truncated backpropagation-through-time \citep{williams1990efficient} on windows of 100 symbols.

Unitary RNNs did not perform well on this task, even after increasing the number of hidden units to $1024$, as shown  in \fig{plot_xml_unitary}.
In general, to remain competitive for online tasks, we believe Unitary RNNs need forget gates.
For the rest of the experiments in this section, we excluded the Unitary RNNs since their high learning curves skewed the plots. 

The remaining XML learning curves are shown in \fig{plot_xml}. Associative LSTM demonstrates a significant advantage that increases with the number of redundant copies.
We also added a comparison to a larger LSTM network with $512$ memory cells.
This has the same number of cells as the Associative LSTM with 4 copies of 128 units, yet Associative LSTM with $4$ copies still learned significantly faster.
Furthermore, Associative LSTM with 1 copy of 128 units greatly outperformed LSTM with 128 units, which appeared to be unable to store enough characters in memory.

We hypothesise that Associative LSTM succeeds at this task by using associative lookup to implement multiples queues to hold tag names from different nesting levels.
It is interesting that $4$ copies were enough to provide a dramatic improvement, even though the task may require up to $40$ characters to be stored (e.g., when nesting $4$ tags with $10$ characters in each tag name).

\subsubsection{Variable Assignment}
\begin{figure}[t]
\begin{center}
\includegraphics[width=\figwidth]{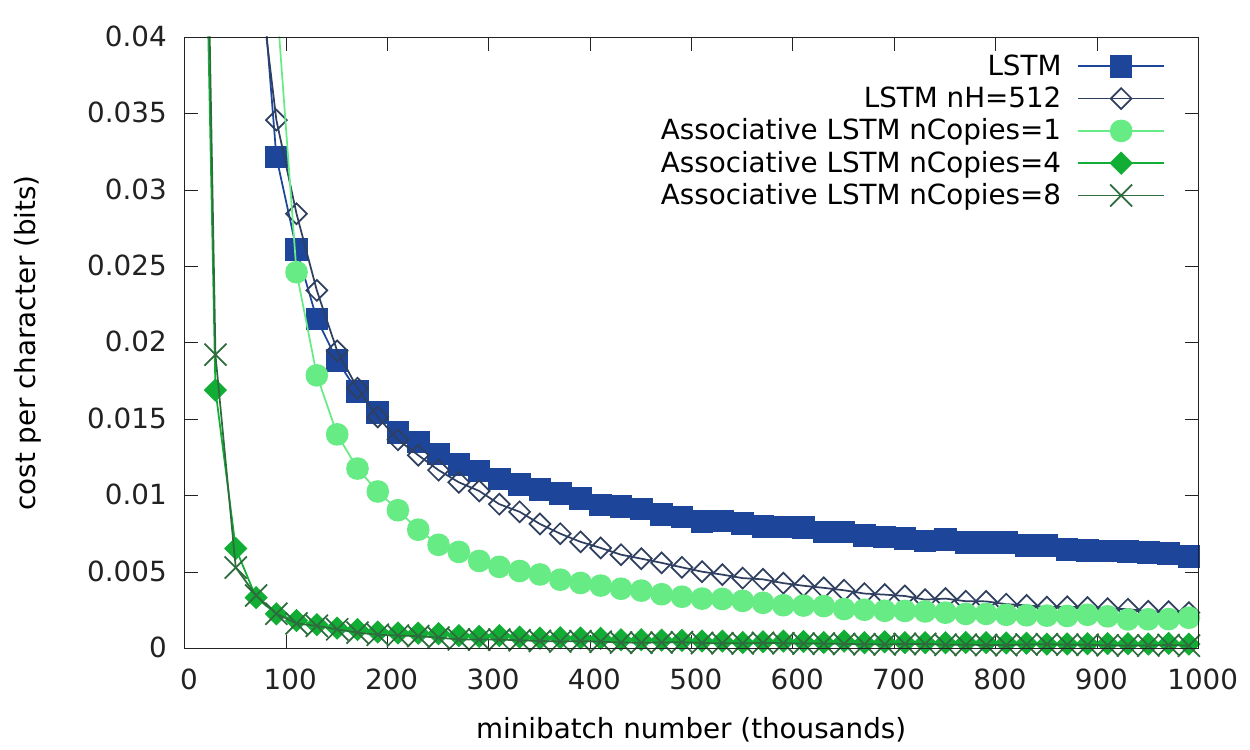}
\end{center}
\caption{Training cost on the variable assignment task.
}
\label{fig:plot_setget}
\figskip
\end{figure}
The variable assignment task was designed to test the network's ability to perform key-value retrieval.
A simple synthetic language was employed, consisting of sequences of 1 to 4 assignments of the form \emph{s(variable,value)}, meaning ``set variable to value'', followed by a query token of the form \emph{q(variable)}, and then the assigned value the network must predict.
An example sequence is the following (prediction targets underlined):

\texttt{s(ml,a),s(qc,n),q(ml)\target{a.}s(ksxm,n),s(u,v)
,s(ikl,c),s(ol,n),q(ikl)\target{c.}s(}\dots

The sequences are presented one character at a time to the networks. The variable names were random strings of 1 to 4 characters, while each value was a single random character.

As shown in \fig{plot_setget}, the task is easily solved by Associative LSTM with 4 or 8 copies, while LSTM with 512 units solves it more slowly, and LSTM with 128 units is again worse than Associative LSTM with a single copy.
Clearly, this is a task where associative recall is beneficial, given Associative LSTM's ability to implement associative arrays.

\subsubsection{Arithmetic}

\begin{table*}[t]
\caption{Network sizes for the arithmetic task.}
\label{tab:netarithmetic}
\begin{center}
\begin{tabular}{l l l r}
\hline
Network & $N_h$ & Relative speed & \#parameters \\
\hline
LSTM & $128$ & 1 & $117,747$ \\
LSTM nH=512 & $512$ & 0.15 & $1,256,691$ \\
\hline
Associative LSTM & $128$ (=$64$ complex numbers) & \begin{tabular}{@{}l@{}}
  0.19 ($N_\mathit{copies}=1$) \\
  0.15 ($N_\mathit{copies}=4$) \\
  0.11 ($N_\mathit{copies}=8$)
\end{tabular} & $131,123$ \\
\hline
Associative LSTM nHeads=3 & $128$ (=$64$ complex numbers) & \begin{tabular}{@{}l@{}}
  0.10 ($N_\mathit{copies}=1$) \\
  0.08 ($N_\mathit{copies}=4$) \\
  0.07 ($N_\mathit{copies}=8$)
\end{tabular} & $775,731$ \\
\hline
\end{tabular}
\end{center}
\end{table*}

\begin{figure}[t]
\begin{center}
\includegraphics[width=\figwidth]{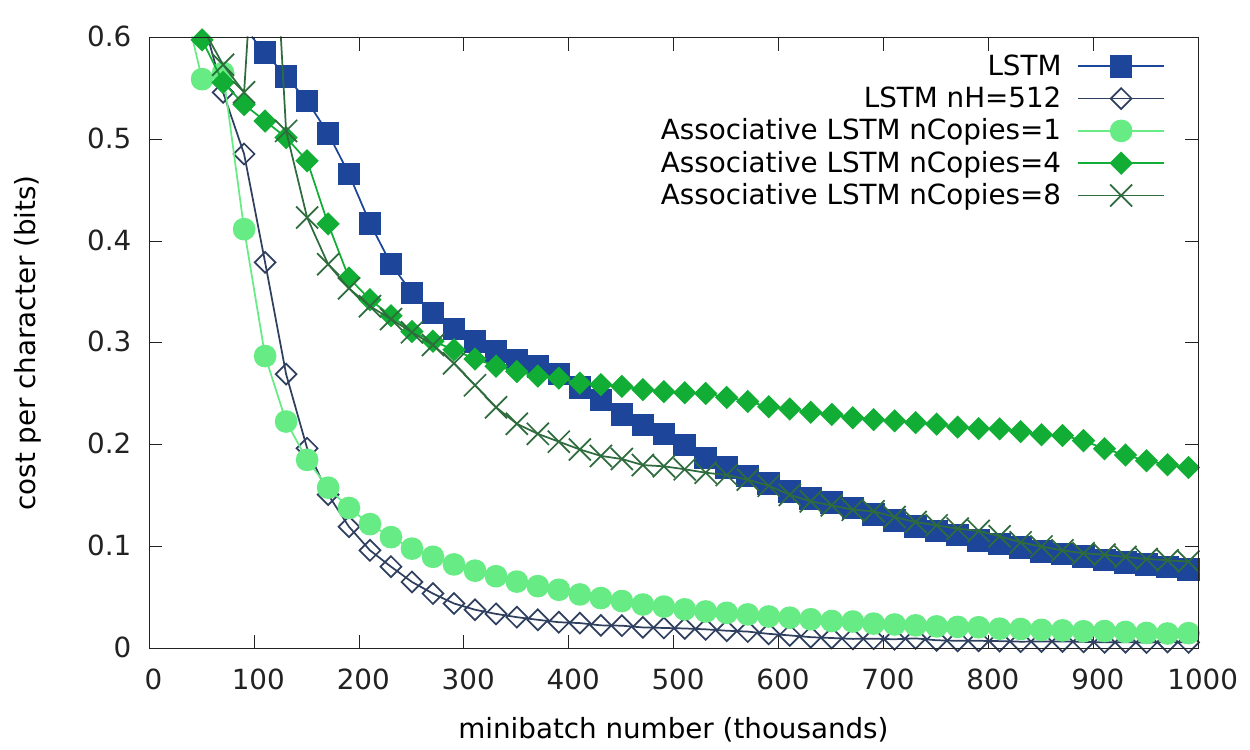}
\end{center}
\caption{Training cost on the arithmetic task.
}
\label{fig:plot_arithmetic}
\end{figure}

\begin{figure}[t]
\begin{center}
\includegraphics[width=\figwidth]{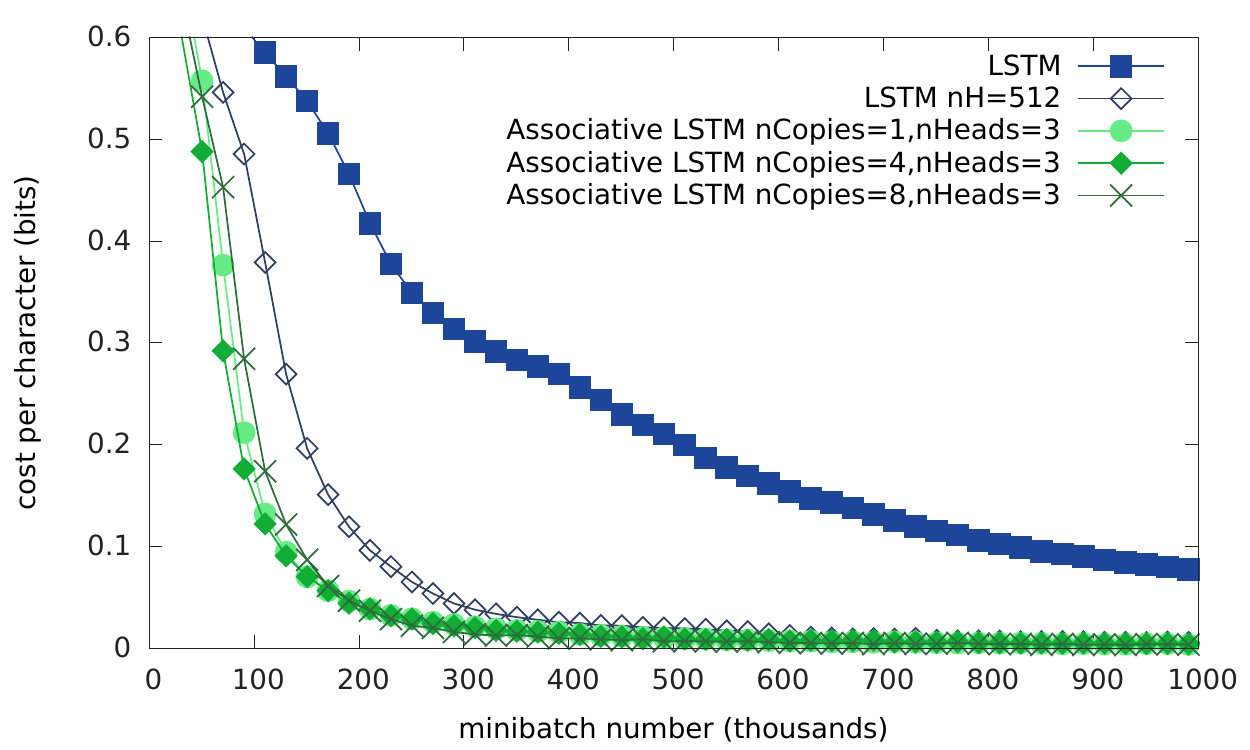}
\end{center}
\caption{Training cost on the arithmetic task
when using Associative LSTM with 3 writing and reading heads.
}
\label{fig:plot_arithmetic_heads}
\figskip
\end{figure}

We also evaluated Associative LSTM on an arithmetic task.
The arithmetic task requires the network to add or subtract two long numbers represented in the input by digit sequences.
A similar task was used by \cite{jozefowicz2015empirical}.

An example sequence is: 

\texttt{-4-98308856=\target{06880389-]}-981+1721=\target{047]}-10
+75723=\target{31757]}8824413+}\dots 

The character ``\texttt{]}'' delimits the targets from a continued sequence of new calculations.
Note that each target, ``$-98308860, \dots$'', is written in reversed order: ``\texttt{\target{06880389-}}''. This trick, also used by \cite{stackrnn} for a binary addition task, enables the networks to produce the answer starting from the least significant digits.
The length of each number is sampled uniformly from 1 to 8 digits.

We allowed the Associative LSTM to learn the $W_{hu}$ matrix in Equation~\ref{eq:whu}, which enables the cell state to compute a carry recurrently. 
Network parameters and execution speeds are shown in Table~\ref{tab:netarithmetic}.

Associative LSTM with multiple copies did not perform well on this task, though the single-copy variant did well (Figure~\ref{fig:plot_arithmetic}). 
There is a subtle reason why this may have occurred. 
Associative LSTM is designed to read from memory only one value at a time, whereas the addition task requires retrieval of three arguments: two input digits and a carry digit. 
When there is only one copy, Associative LSTM can learn to write the three arguments to different positions of the hidden state vector. But when there are multiple copies, the permutation matrices exclude this solution.  
This is not an ideal result as it requires trading off capacity with flexibility, but we can easily construct a general solution by giving Associative LSTM the ability to read using multiple keys in parallel.
Thus, we built an Associative LSTM with the ability to write and read multiple items in one time step simply by producing $N_\mathit{heads}$ input and output keys. 

As shown in \fig{plot_arithmetic_heads}, adding multiple heads helped as Associative LSTM with 3 heads and multiple copies consistently solved the task.

\subsubsection{Wikipedia}

\begin{figure}[t]
\begin{center}
\includegraphics[width=\figwidth]{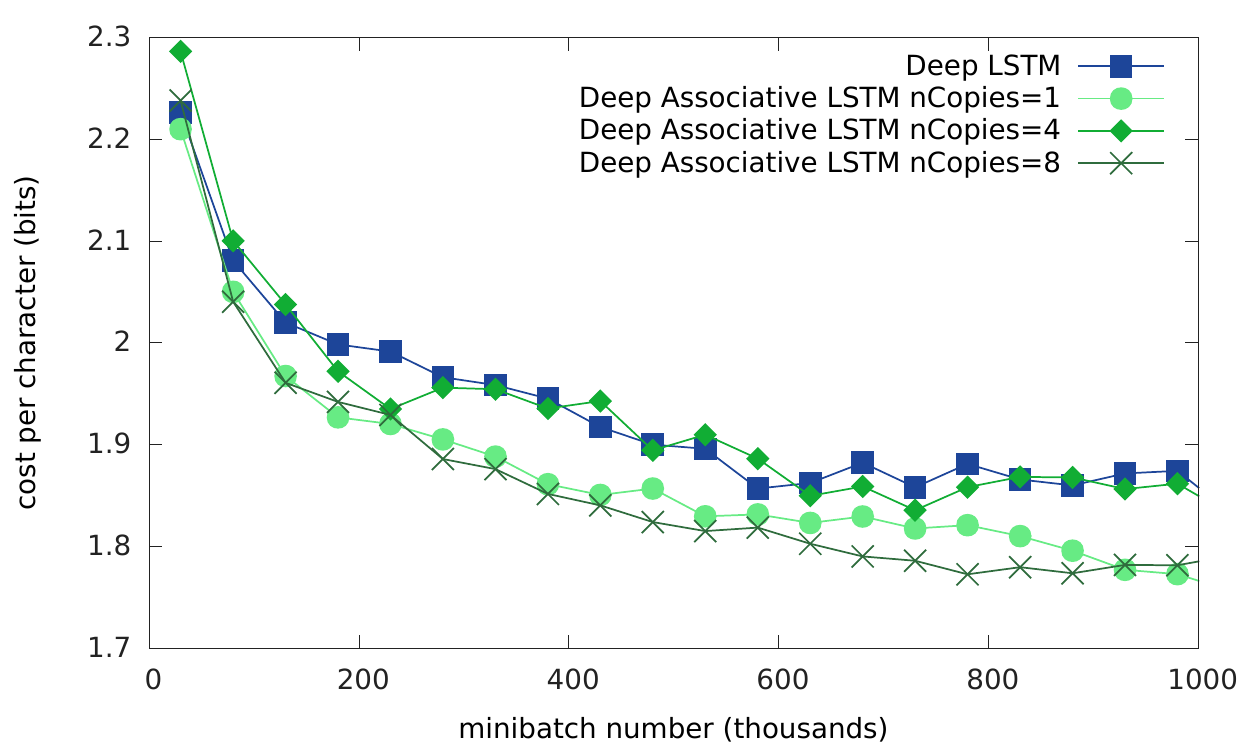}
\end{center}
\caption{Test cost when modeling English Wikipedia.
}
\label{fig:plot_enwik}
\figskip
\end{figure}

The last task is the sequence prediction of English Wikipedia \citep{huttercontest}, which we used to test whether Associative LSTM is suitable for a natural language processing task.

The English Wikipedia dataset has 100 million characters of which we used the last 4 million as a test set.
We used larger models on this task but, to reduce training time, did not use extremely large models since our primary motivation was to compare the architectures.
LSTM and Associative LSTM were constructed from a stack of 3 recurrent layers with $256$ units by sending the output of each layer into the input of the next layer as in \cite{graves2013speech}.

Associative LSTM performed comparably to LSTM (\fig{plot_enwik}). We expected that Associative LSTM would perform at least as well as LSTM; if the input key $r_i$ is set to the all $1$-s vector, then the update in Equation~\ref{eq:cellupdate} becomes 
\begin{align}
  c_{s,t} &= g_f \elemmul c_{s,t-1} + g_i \elemmul u
\end{align}
which exactly reproduces the cell update for a conventional LSTM. Thus, Associative LSTM is at least as general as LSTM.

\section{Why use complex numbers?}
We used complex-valued vectors as the keys.
Alternatively, the key could be represented by a matrix $A$, and the complex multiplication then replaced with matrix multiplication $A x$. To retrieve a value associated with this key, the trace would be premultiplied by $A^{-1}$. 
Although possible, this is slow in general and potentially numerically unstable.

\section{Conclusion}
Redundant associative memory can serve as a new neural network building block. 
Incorporating the redundant associative memory into a recurrent architecture with multiple read-write heads provides flexible associative storage and retrieval, high capacity, and parallel memory access. 
Notably, the capacity of Associative LSTM is larger than that of LSTM without introducing larger weight matrices, and the update equations of Associative LSTM can exactly emulate LSTM, indicating that it is a more general architecture, and therefore usable wherever LSTM is. 

\section*{Acknowledgments}
We would like to thank Edward Grefenstette, Razvan Pascanu, Timothy Lillicrap, Daan Wierstra and Charles Blundell for helpful discussions.

\bibliography{complexlstm}
\bibliographystyle{icml2016}
}

\appendix

\section{Comparison with a Neural Turing Machine}
We have run additional experiments to compare Associative LSTM with a Neural Turing Machine (NTM). The comparison was done on the XML, variable assignment and arithmetic tasks. The same network architecture was used for all three tasks.
The network sizes can be seen in Table~\ref{tab:netntm}.

\begin{figure}[H]
\begin{center}
\includegraphics[width=\figwidth]{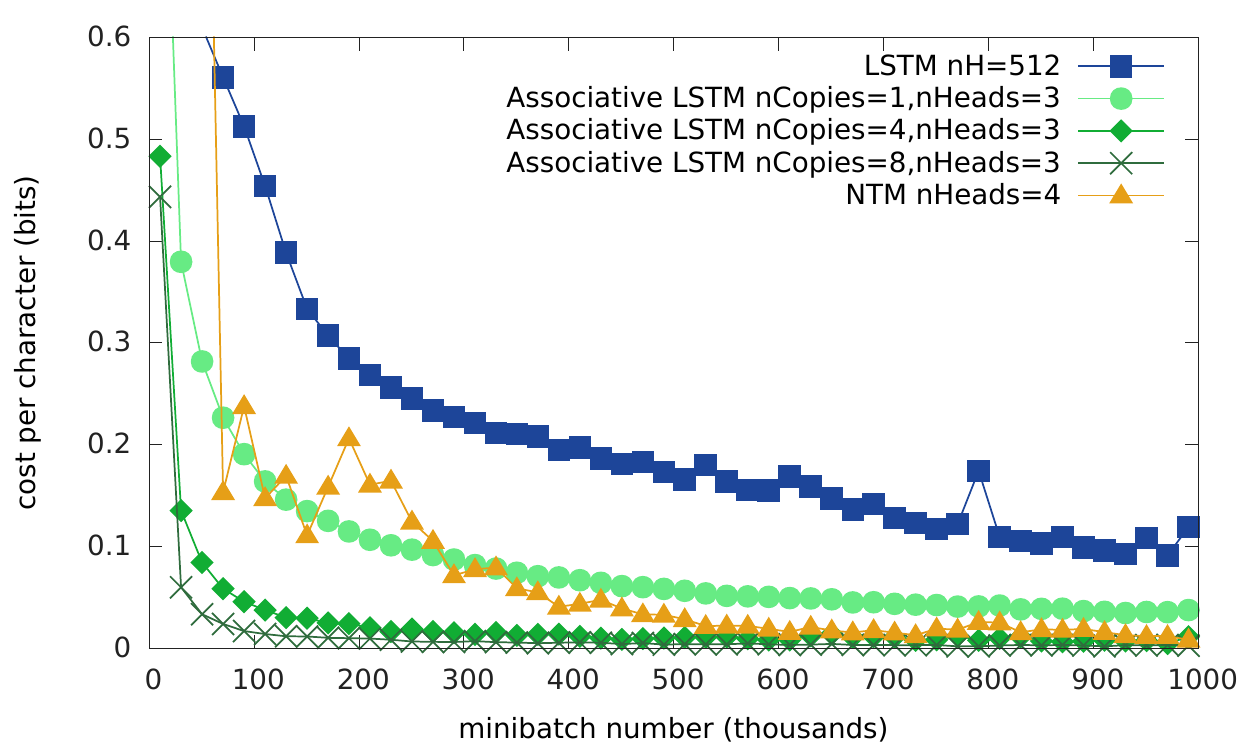}
\end{center}
\caption{Training cost on the XML task.
}
\label{fig:plot_xml_ntm}
\figskip
\end{figure}

\begin{figure}[H]
\begin{center}
\includegraphics[width=\figwidth]{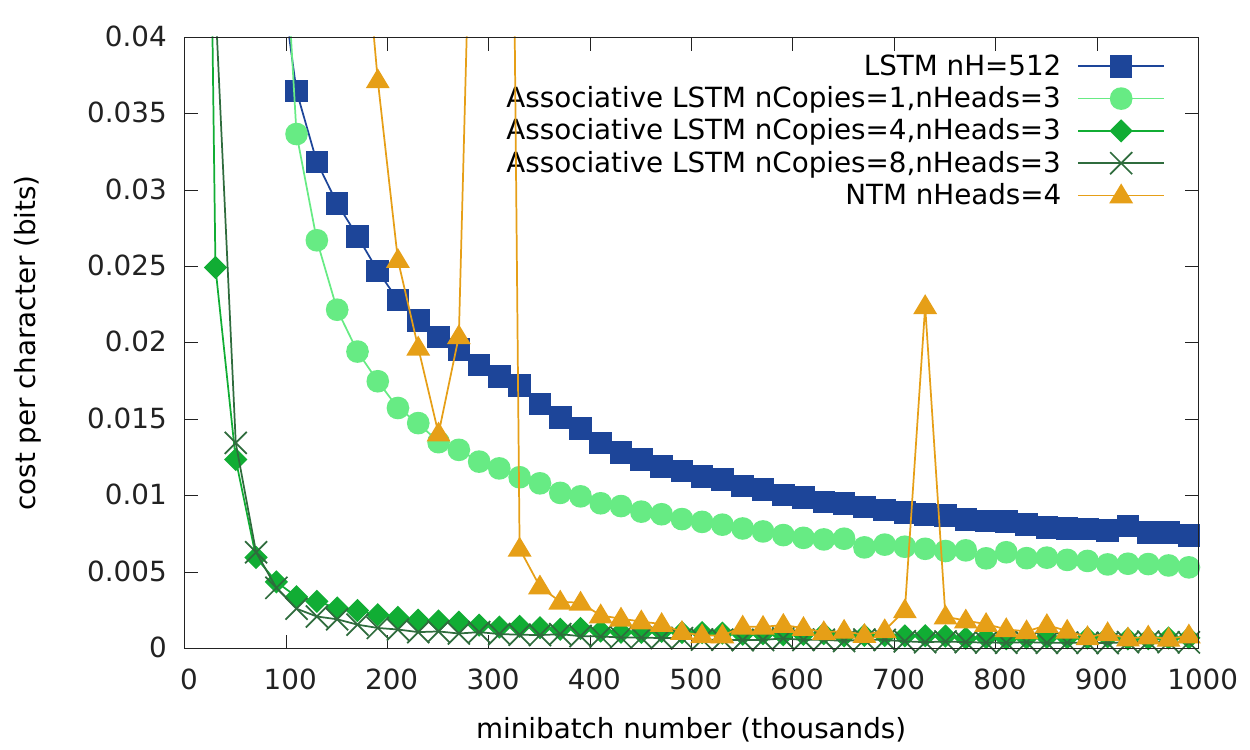}
\end{center}
\caption{Training cost on the variable assignment task.
}
\label{fig:plot_setget_ntm}
\figskip
\end{figure}

\begin{figure}
\vskip -1.62cm
\begin{center}
\includegraphics[width=\figwidth]{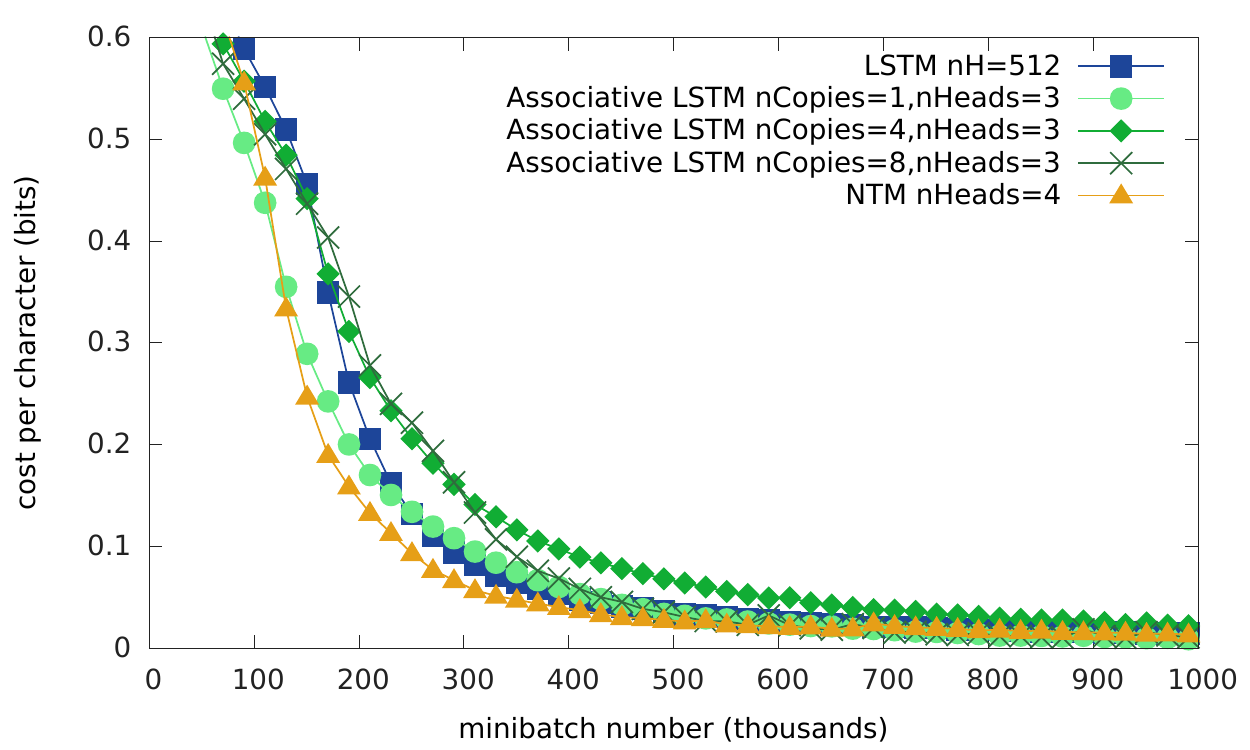}
\end{center}
\caption{Training cost on the arithmetic task.
}
\label{fig:plot_arithmetic_ntm}
\figskip
\end{figure}

The learning curves are shown in Figures~\ref{fig:plot_xml_ntm}-\ref{fig:plot_arithmetic_ntm}. Training was done with minibatches of size 1 to be able to compare with the original Neural Turing Machine, but other minibatch sizes lead to similar learning curves.
Both Associative LSTM and the Neural Turing Machine achieved good performance on the given tasks. Associative LSTM has more stable learning progress.
On the other hand, the Neural Turing Machine has shown previously better generalization to longer sequences on algorithmic tasks.

\captionof{table}{Networks compared to a Neural Turing Machine.}
\vskip -0.7cm
\label{tab:netntm}
\begin{center}
\begin{tabular}{l l l r}
\hline
Network & Memory Size & Relative speed & \#parameters \\
\hline
LSTM nH=512 & $N_h=512$ & 1 & $1,256,691$ \\
\hline
Associative LSTM nHeads=3 & $N_h=128$ (=$64$ complex numbers) & \begin{tabular}{@{}l@{}}
  0.66 ($N_\mathit{copies}=1$) \\
  0.56 ($N_\mathit{copies}=4$) \\
  0.46 ($N_\mathit{copies}=8$)
\end{tabular} & $775,731$ \\
\hline
NTM nHeads=4 & $N_h=384, \mathrm{memorySize}=128 \times 20$ & 0.66 & $1,097,340$ \\
\end{tabular}
\end{center}

\end{document}